\pdfoutput=1

\documentclass[11pt]{article}

\usepackage[preprint]{acl}

\usepackage{times}
\usepackage{latexsym}

\usepackage[T1]{fontenc}
\usepackage[utf8]{inputenc}
\usepackage{microtype}
\usepackage{inconsolata}
\usepackage{graphicx}
\usepackage{amsmath}
\usepackage{booktabs}
\usepackage{multirow}
\usepackage{tikz}
\usetikzlibrary{arrows.meta,positioning}

\title{Structure for Reading, Prose for Writing: Asymmetric Structural\\Conditioning in Multi-Agent Document Authoring}

\author{Cheng Yu\textsuperscript{1}, Nikhil Mathew\textsuperscript{1}, Zhengjie Wang\textsuperscript{1} \\
  \textsuperscript{1}ML Research Labs\thanks{This work was funded by ML Research Labs (a Trellis Data company).}, Canberra, Australia
\\
  \small{
    \textbf{Correspondence:} \href{mailto:cheng.yu@mllabs.com.au}{cheng.yu@mllabs.com.au}
  }
}

\begin{document}
\maketitle
\begin{abstract}
Multi-agent pipelines that author formal documents must both read a requester's forms and write against them. We report a deployed tender-response system, running an open-weights model under sovereignty constraints, and evaluate it against human-written bids the same organisation actually submitted. On a blind comparison where the system had no worked example available, an LLM judge rated its answers at least as good as the human-submitted answer on $40$ of $55$ ground-truth sections, better on $4$, missing on none, and flagged one unsupported claim in total. Classifying every gap the judge identified shows that $68\%$ were content absent from the system's own sources -- knowledge the human author held and the pipeline was never given -- so only $6$ of the $15$ adverse verdicts involve a deficiency the system could have avoided. A divergence from ground truth is more often an information-availability result than a writing-quality one, and evaluations that do not separate the two understate such systems. Against this backdrop we report a conditioning asymmetry. It is well established that rendering documents as structural markup rather than flat prose improves extraction, and we reproduce that on three reading tasks. The benefit does not transfer to conditioning: converting a bid's \emph{instruction} material from prose to nested XML dropped answer quality from $74\%$ to $48\%$ under a paired comparison. We further find that naming a forbidden construction concentrates rather than removes it -- $96\%$ of surviving defects fall in the two forms the prompt explicitly names -- and that coupling a stochastic annotation to a deterministic windowing function moves the extracted requirement count from $68$ to $51$ on a byte-identical file. Structure belongs where the model reads; prose and self-applied tests belong where it writes.
\end{abstract}

\begin{figure*}[t]
    \centering
    \includegraphics[width=0.98\textwidth]{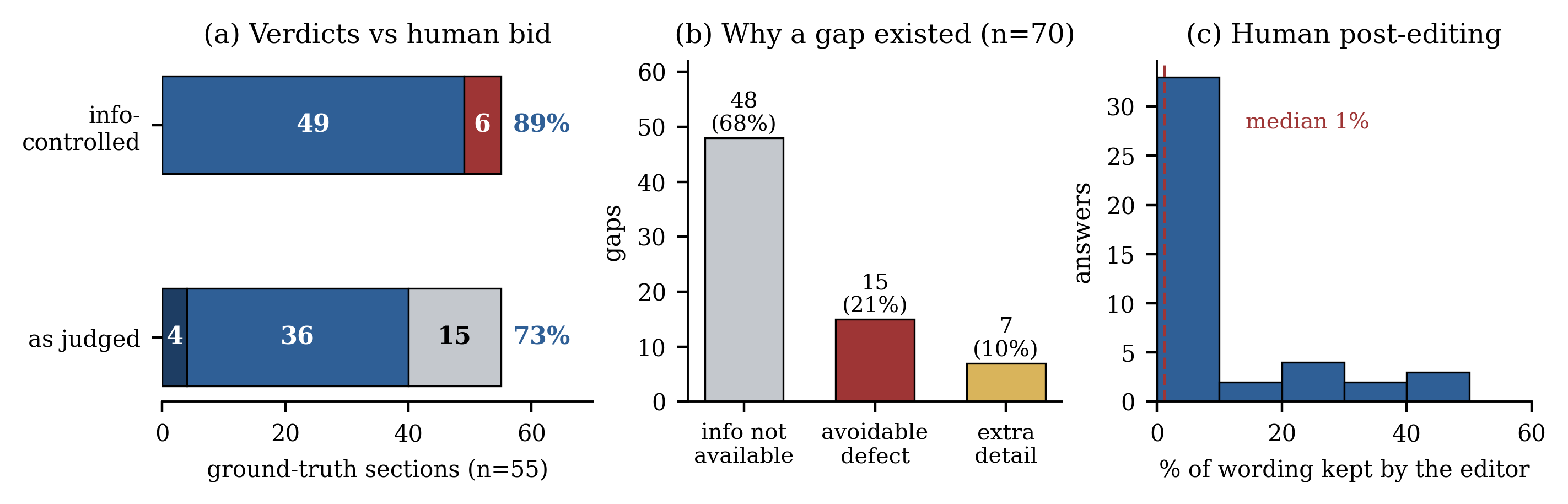}
    \caption{Evaluation against human work. \textbf{(a)} The system's answers judged against the human-written bid the organisation submitted for the same procurement, by an LLM instructed to score on merit rather than similarity; the system had no exemplar available for this tender. The lower bar re-scores after excluding adverse verdicts driven only by information the system's sources did not contain (\S\ref{sec:truth}). \textbf{(b)} Every gap the judge identified, classified by cause. \textbf{(c)} On a second procurement a human editor worked \emph{from} the system's draft; the histogram is the share of each drafted answer's wording that survived into the submitted document (\S\ref{sec:postedit}).}
    \label{fig:truth}
\end{figure*}

\section{Introduction}

Formal document authoring -- tender responses, regulatory filings, compliance matrices -- inverts the usual generative setting. The output must be returned inside the requester's own files, in their layout; every obligation in a scattered document set must be answered; and the requester's wording must survive verbatim, because a paraphrased requirement is one an evaluator will not recognise. Under data-sovereignty constraints the model doing this work cannot be a frontier API, which removes the headroom other systems use to absorb document complexity. Prior work in this domain establishes that an isolated model is insufficient for it: reliable extraction from heterogeneous artefacts, consistent multi-step analysis and systematically validated output require the reading, extracting and drafting roles to be separated \cite{hendrata2026tendering}.

Multi-agent decomposition is the standard response. Co-Scientist \cite{gottweis2026coscientist} demonstrates that isolating generation from critique across specialised agents materially improves reliability, and directed acyclic graphs have become the orchestration substrate for such systems \cite{sdag2026}, replacing open-ended conversational loops that drift. A parallel literature establishes that language models read structure better than prose: constraining generation to a verifiable extract--validate--enumerate pipeline raises F1 by $31\%$ \cite{chen2026eve}, serialisation format alone is worth up to $8.8$ points of accuracy on questionnaire understanding \cite{nguyen2025qasu,sui2024table}, and supplying hierarchical control-flow structure instead of flattened decompiler output raises compilability from $45.0\%$ to $85.2\%$ \cite{achamyeleh2026helios}. A third literature examines how errors move through sequential agent chains, and does not agree on the sign: detection of an injected falsehood falls from $72.0\%$ to $50.9\%$ across four agents \cite{snowball2026}, while a study over $500$ cascades reports deeper chains \emph{lowering} aggregate hallucination at the cost of factual accuracy \cite{jamshidi2026cascade}. Both motivate verification at every handoff \cite{lin2025agentask,xie2026spark} and aggressive context compaction \cite{kang2025acon}.

Every one of these structural results is measured on a \emph{reading} task -- find the value, pair the instruction, enumerate the fields. We are not aware of work that tests whether the same benefit holds when the structured material is not the document to be read but the instructions the model must internalise and write from. This paper reports that it does not.

Our contributions are as follows.

\begin{itemize}
\itemsep-0.15em
\item An evaluation of a deployed system against human bids the same organisation submitted, including one blind comparison in which the system had no worked example (\S\ref{sec:truth}), and a post-editing measurement on a second procurement where a human worked from the machine draft (\S\ref{sec:postedit}).
\item A method for reading such comparisons correctly: classifying each gap by whether the substance was available to the system at all separates information-availability failures from writing failures, and moves the score from $73\%$ to $89\%$ (\S\ref{sec:truth}).
\item A controlled paired comparison showing that structural markup, which helps on every reading task we instrumented (\S\ref{sec:reading}), \emph{reverses} when applied to instruction material (\S\ref{sec:writing}).
\item A surface-form-resolved measurement of prohibition naming: the residual defects concentrate in exactly the forms the prompt names (\S\ref{sec:prohibition}).
\item A variance-propagation mechanism absent from the error-cascade literature, in which a deterministic windowing function fed a stochastic annotation amplifies upstream variance (\S\ref{sec:variance}).
\end{itemize}

\section{System}
\label{sec:system}

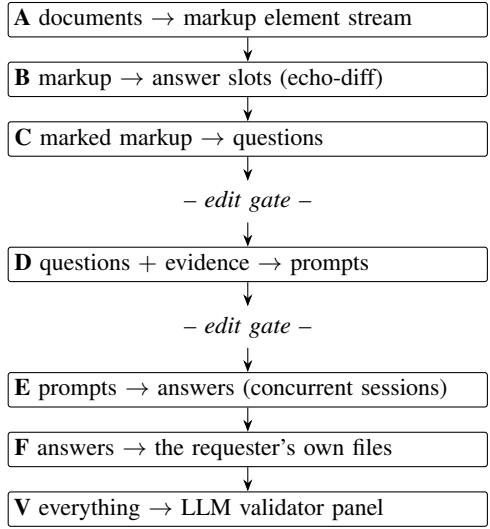
\begin{figure}[h]
    \centering
    \begin{tikzpicture}[
      node distance=3.2mm,
      box/.style={draw, rounded corners=1pt, inner sep=2.4pt, font=\small,
                  text width=0.80\linewidth, align=left},
      gate/.style={font=\footnotesize\itshape, text width=0.80\linewidth, align=center},
      >={Stealth[length=1.6mm]}]
      \node[box] (a) {\textbf{A} documents $\rightarrow$ markup element stream};
      \node[box, below=of a] (b) {\textbf{B} markup $\rightarrow$ answer slots (echo-diff)};
      \node[box, below=of b] (c) {\textbf{C} marked markup $\rightarrow$ questions};
      \node[gate, below=of c] (g1) {-- edit gate --};
      \node[box, below=of g1] (d) {\textbf{D} questions $+$ evidence $\rightarrow$ prompts};
      \node[gate, below=of d] (g2) {-- edit gate --};
      \node[box, below=of g2] (e) {\textbf{E} prompts $\rightarrow$ answers (concurrent sessions)};
      \node[box, below=of e] (f) {\textbf{F} answers $\rightarrow$ the requester's own files};
      \node[box, below=of f] (v) {\textbf{V} everything $\rightarrow$ LLM validator panel};
      \draw[->] (a) -- (b); \draw[->] (b) -- (c); \draw[->] (c) -- (g1);
      \draw[->] (g1) -- (d); \draw[->] (d) -- (g2); \draw[->] (g2) -- (e);
      \draw[->] (e) -- (f); \draw[->] (f) -- (v);
    \end{tikzpicture}
    \caption{The six-stage graph. Stage V runs on every execution, non-fatally.}
    \label{fig:pipeline}
\end{figure}

The system is a directed graph of $43$ single-shot agent roles plus one multi-turn drafter, each with its own system prompt and each receiving only the context its task requires. It runs in six stages across three human edit gates, so expensive semantic work is done once, persisted as inspectable JSON, and can be hand-corrected before the next stage consumes it (Figure~\ref{fig:pipeline}).

\paragraph{Markup and locators.} Every input document is parsed into an ordered element stream -- one paragraph, table cell, spreadsheet cell or slide shape per element -- carrying a synthetic identifier \texttt{eid}, a reversible \texttt{loc} into the source file, the rendered text, and format flags. The stream is rendered as nested markup that preserves the document tree, because a flat one-line-per-element rendering destroys the containment and adjacency that pair a question with its instruction and its answer box:

\vspace{0.6em}
\noindent\begin{minipage}{\linewidth}
{\small\begin{verbatim}
<table id="t3" heading="Requirement 2">
 <row n="7">
  <cell n="0" eid="E0184">2.1 Describe...
  <cell n="1" eid="E0185">
   <note>Max 2 pages.</note>
  <cell n="2" eid="E0186" answer_slot="yes">
\end{verbatim}}
\end{minipage}
\vspace{0.4em}

\noindent Two rules govern the whole system. \textbf{Locators address; they never classify}: the model decides what an element \emph{is} from structure it can see, and a coordinate is used only afterward, to write an answer back. \textbf{Requirement text is spliced by code} from the element stream and never retyped by a model. Together these mean a hallucinated coordinate cannot reach the file writer and a paraphrased requirement cannot reach the evaluator.

\paragraph{Answer slots by echo-diff.} Rather than describing slots in a schema, we hand the model one markup window and ask it to echo the window back unchanged except that each cell a respondent must type into becomes a sentinel token; code diffs echo against input. The model performs one transformation instead of a twelve-field classification, and an \texttt{eid} it never received cannot appear in its output.

Window size is tuned against reproduction fidelity, which decays \emph{mid}-window rather than at the edges. On the same document, a $23.5$k-character window silently dropped marks at relative position $0.61$--$0.63$ and found $30$ of $32$ slots; an $8$k window echoed all $407$ elements and found $32$ of $32$, $5.7\times$ faster, because short calls parallelise where long ones serialise on generation. Because the loss is silent and mid-window, the pass counts elements that entered a window and did not return, rather than assuming none were lost. Enumeration wants the opposite setting: question extraction runs over larger windows, since breadth lets the model see that an obligation stated two questions away belongs here. Chunk size is therefore a per-task parameter, not a system constant.

\paragraph{Drafting sessions.} A section is one chat session and the unit of parallelism; questions within it are serial, because the accumulating history is what enforces consistent terminology and figures across related answers. Sessions are scheduled continuously on first-completion rather than in synchronised levels, since chains are of very unequal length. Each question is authored in four turns: a self-ask plan; a draft conditioned on the closest matching answer from a past submitted bid; a compliance pass against the tender's verbatim requirements; and a quality pass that may add, reframe or flag but may never delete.

The exemplar is retrieved by matching the incoming requirement against the \emph{question} a past requester asked rather than against our own prose, which is the offline question-to-question symmetry of \citet{vake2026hype} with the hypothetical step removed -- in a corpus of bids we submitted, the past questions are recovered rather than synthesised. A tender never sees its own past response: self-exclusion is enforced, and the same-product gate is hard, so with a small library the common outcome is no exemplar at all. That path must therefore be silent and harmless, and \S\ref{sec:truth} exercises it.

\section{Experiments}
\label{sec:experiments}

\subsection{Setup}

We evaluate on four public-sector procurements in Australia and New Zealand, denoted T1--T4. Requester identities, sectors and document content are withheld: these are competitive commercial bids. Each exists in two forms -- the response the organisation actually wrote and submitted, and the response this system generated for the same blank forms -- which is what makes \S\ref{sec:truth} and \S\ref{sec:postedit} possible.

Generation runs on an open-weights model of under 200B parameters, served from Australian infrastructure, with no thinking tokens and no server-side conversation state, across two endpoints under independent AIMD rate limiters. It is not a frontier model, which matters for \S\ref{sec:prohibition}. \textbf{Temperature is left at the model default throughout}, deliberately: reproducibility claims must hold in the configuration that is actually deployed rather than at temperature zero.

All quality verdicts come from LLM validator agents; we use no lexical rules for scoring. The verdict of record for answer quality is a three-way judgement per answer -- \emph{answering}, \emph{partly restating}, or \emph{mostly restating}. Where we quote a lexical count it was computed outside the pipeline as corroboration only, and \S\ref{sec:corroboration} reports why that corroboration is unreliable.

\subsection{Against human ground truth}
\label{sec:truth}

For T3 the organisation submitted a human-written response before this system existed. That response is therefore usable as blind ground truth, and because self-exclusion applies, the system could not condition on it: \textbf{$0$ of $120$ T3 prompts carried an exemplar block}. The system wrote T3 with no worked example of any kind.

We recovered the human submission as $55$ (question, answer) sections and asked a validator to compare each against whatever combination of our sections covers the same ground. The judge is instructed to score on \emph{merit, not similarity} -- responsiveness, substance, discipline and depth -- and told explicitly that the ground truth is a good answer, not the only good answer. It emits \emph{better}, \emph{as\_good}, \emph{weaker} or \emph{missing}, plus any claims of ours that look invented.

Results are in Figure~\ref{fig:truth}(a) and Table~\ref{tab:truth}. The system was rated at least as good on $40$ of $55$ sections ($73\%$), better on $4$, and \emph{missing} on none -- every ground-truth section was covered by something. Across all $55$ sections the judge flagged \textbf{one} unsupported claim, an inference of multi-factor authentication from a passage that described only single sign-on.

\begin{table}[h]
  \centering
  \caption{T3: system answers judged against the human-submitted bid, $n{=}55$ ground-truth sections.}
  \small
  \begin{tabular}{lcr}
    \toprule
    Verdict & $n$ & Share \\
    \midrule
    better & 4 & 7\% \\
    as good & 36 & 65\% \\
    weaker & 15 & 27\% \\
    missing & 0 & 0\% \\
    \midrule
    \textbf{at least as good} & \textbf{40} & \textbf{73\%} \\
    \midrule
    sections w/ unsupported claim & 1 & 2\% \\
    \bottomrule
  \end{tabular}
  \label{tab:truth}
\end{table}

\paragraph{A divergence is not automatically a defect.} The $15$ adverse verdicts carry $70$ specific gaps, and taking them at face value as quality failures would be a category error. A generated answer can only contain what the system's sources support. Where the human author knew something from experience, a colleague or a conversation that never entered the pipeline, the resulting divergence measures information availability, not writing.

We therefore classified every gap. For each, we assembled the exact corpus the system had for this tender -- the eight reference sources at their configured caps plus the tender's own documents and four addenda, $372{,}139$ characters in total -- recorded which named entities in the gap occur anywhere in it, and asked a classifier to assign one of four causes: the substance was \emph{unavailable} in our sources; it was \emph{available and unused}; it is \emph{policy divergence} (something the organisation has since stopped claiming); or it is \emph{additional detail} the question did not require.

\begin{table}[h]
  \centering
  \caption{Cause of every gap the judge identified on T3 ($n{=}70$ after discarding two empty records).}
  \small
  \begin{tabular}{lcr}
    \toprule
    Cause & $n$ & Share \\
    \midrule
    Information unavailable to the system & 48 & 68\% \\
    Additional, non-required detail & 7 & 10\% \\
    \midrule
    \textbf{Avoidable -- available but unused} & \textbf{15} & \textbf{21\%} \\
    \bottomrule
  \end{tabular}
  \label{tab:gaps}
\end{table}

Table~\ref{tab:gaps} shows that $68\%$ of gaps were content the pipeline never had: a named integration platform, a curated library of role-specific prompts, a request the organisation had made to the requester about confidentiality, the ABNs of subcontractors recorded nowhere in its inputs. A further $10\%$ are supplementary colour that does not affect responsiveness. Only $21\%$ are avoidable -- present in the sources and not used -- and those are the ones worth engineering against.

Re-scoring on that basis, $9$ of the $15$ adverse verdicts are driven \emph{only} by unavailable information or optional detail, leaving $6$ that involve any avoidable gap. The information-controlled figure is therefore $49$ of $55$ ($89\%$), against $40$ of $55$ ($73\%$) as judged (Figure~\ref{fig:truth}a). Neither number is the whole truth: $73\%$ understates the writing by charging it for missing knowledge, and $89\%$ flatters the deployed system, because a bid that omits a subcontractor's ABN is still deficient as a submission whatever the cause. The pair brackets the result, and the gap between them is the value of better information plumbing rather than better generation.

One gap deserves separate mention because it inverts entirely. An adverse verdict records that our answer "fails to list the multiple subcontractors\ldots with ABNs/ACNs as in the ground truth". One of those entities is one the organisation has since stopped naming, and a standing rule in the system's always-binding block forbids naming it. The system was penalised for following ground truth \emph{newer} than the document it was judged against. Any evaluation against a past human artefact carries some share of this, and it places a ceiling on the achievable score that is not a ceiling on quality.

We do not claim the system writes better bids than the team. We claim that on a blind section-by-section comparison, with no exemplar, its answers were judged competitive with submitted human work on roughly three quarters of the document; that four fifths of the shortfall traces to information it was never given rather than to how it wrote; and that its claim discipline held at one flagged inference across $55$ sections.

\subsection{Human post-editing}
\label{sec:postedit}

T4 is the complementary case and must not be confused with the first. There a human worked \emph{from} the system's draft, so the submitted document is a post-edit rather than independent truth. Using it as ground truth would inflate the system's score, so we report it as what it is: a measurement of how much of the machine's work a professional kept.

Naive document comparison is misleading here. The filled human document shares $26\%$ of its $10$-grams with the \emph{blank} form -- the requester's own questions and instructions, present in both. Excluding that boilerplate, $20.5\%$ of the human's own answer text is shared with the system's, and $7.9\%$ of the system's answer text survived into the submission. The shared strings are distinctive and technical ("monitoring is handled through standard opentelemetry and syslog exporters", "seat licence at AUD 24 per seat per month"), so this is reuse, not coincidence.

Per answer (Figure~\ref{fig:truth}(b)), retention is heavily skewed: median $1.2\%$, mean $8.7\%$, no answer above $50\%$, and $9$ of $44$ between $20\%$ and $50\%$. Every pricing answer was retained at $0\%$ -- unsurprising, since those are precisely the cells where the system emits a placeholder for a human decision rather than inventing a number.

The honest reading is that the draft functioned as scaffolding rather than as copy. A fifth of the final human text originated in it, concentrated in technical description, while commercial and pricing language was rewritten wholesale. That is a useful outcome for a drafting aid and a poor one for an autonomous author, and the two readings should not be conflated.

\begin{figure*}[t]
    \centering
    \includegraphics[width=0.98\textwidth]{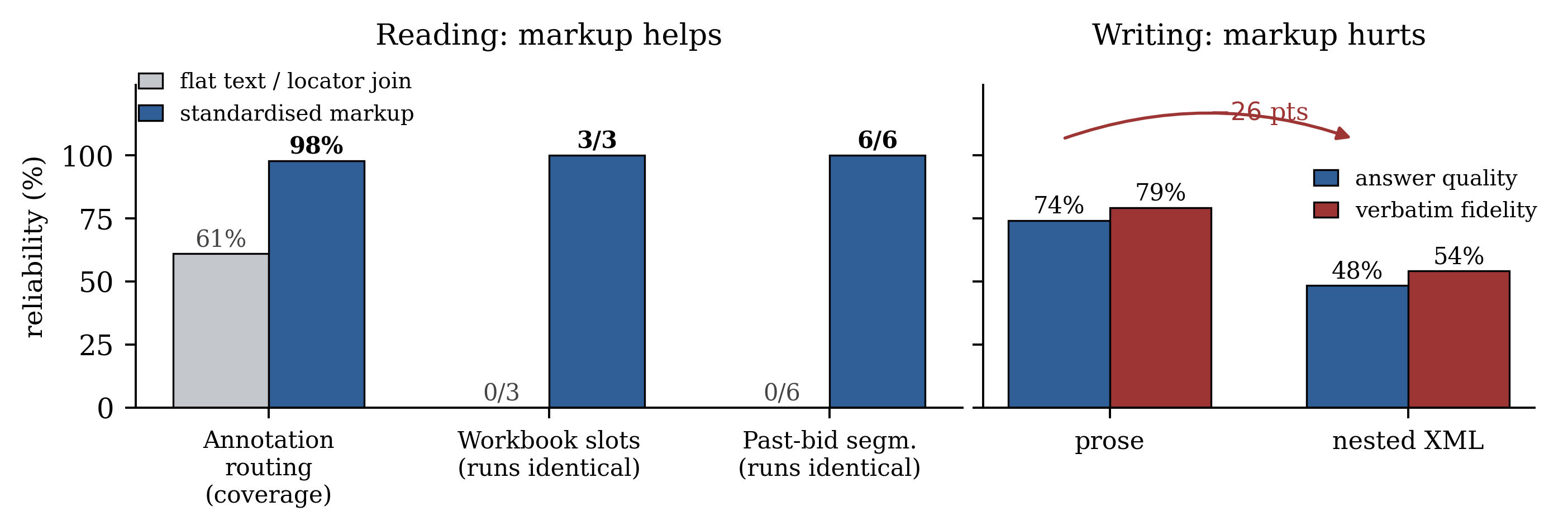}
    \caption{The conditioning asymmetry. \textbf{Left}: three tasks where the model \emph{reads} a document, before and after rendering it as standardised markup -- annotation-to-question routing coverage, and run-to-run identity of extracted workbook answer slots ($3$ runs) and past-bid segmentation ($6$ runs). \textbf{Right}: one task where the model \emph{writes} from instruction material, with that material as prose and as nested XML ($n{=}31$ per arm). Every reading task improved; the writing task regressed on both measures.}
    \label{fig:results}
\end{figure*}

\subsection{Markup on reading tasks}
\label{sec:reading}

We instrumented three reading tasks (Table~\ref{tab:reading}, Figure~\ref{fig:results} left).

Routing team annotations -- margin comments and tracked changes from a previous form revision -- to the questions they govern was originally an exact locator join. A colleague writes beside whatever they are reading, so the annotated element is systematically not one of the question's own elements: $90$ of $228$ annotations ($39\%$) reached no question, silently, and one binding integration requirement consequently appeared in $0$ of $253$ drafted answers. Replacing the join with a batched semantic routing decision -- the model given the annotation, what it was written beside, and the rendered markup six elements before and four after its anchor, with the locator supplied only as a hint -- raised coverage to $97.8\%$ and that requirement to $24$ answers.

Deriving spreadsheet answer cells from a per-cell echo-diff alone returned $404$, $280$ and $340$ slots on three runs of a byte-identical file: how much of the bid got priced was luck. A workbook's shape is a single semantic judgement, so we read it once per tab, verify it with a second adversarial call, and derive the expected answer cells arithmetically. Cross-checking the two views -- keeping the union and logging disagreements rather than letting either win -- gave $310$ slots on three consecutive runs with a residual disagreement of $1$--$2$ cells. Where two model-derived views disagree, that is information rather than a tie for code to break; earlier versions trusted each view alone and each failed, in opposite directions.

Reading a past bid back as (question, answer) pairs from flat text yielded $0$, $1$, $3$, $13$, $18$, $24$ and $28$ pairs across runs. Routing the same documents through the markup path made document segmentation exactly reproducible -- identical slice counts on six runs spanning four days -- and raised the yield floor from $3$ to $47$, though per-pair extraction remains stochastic at a coefficient of variation near $10\%$.

\begin{table}[h]
  \centering
  \caption{Markup on reading tasks. Every task improved.}
  \small
  \begin{tabular}{lcc}
    \toprule
    Task & Flat / join & Markup \\
    \midrule
    Annotation routing & 61\% & \textbf{97.8\%} \\
    Workbook slots ($\times 3$) & 404/280/340 & \textbf{310/310/310} \\
    Past-bid segmentation & varied & \textbf{identical ($\times 6$)} \\
    Past-bid pair floor & 3 & \textbf{47} \\
    \bottomrule
  \end{tabular}
  \label{tab:reading}
\end{table}

\subsection{Markup on a conditioning task}
\label{sec:writing}

We then converted the one piece of \emph{instruction} material in the prompt to the same representation. A per-bid notes file carries the positioning, product decision and overrides for a given tender, and is the highest-precedence source in a $155$k-character reference block. We ran it prose and as nested XML -- \texttt{<bid>} containing \texttt{<requester>}, \texttt{<win\_theme>}, \texttt{<partners>}, \texttt{<capabilities>}, \texttt{<security>}, \texttt{<pricing>}, with attributes for entity identifiers, roles and rates -- holding tender, forms, sources, pipeline and sample size ($n{=}31$) fixed.

Quality fell (Table~\ref{tab:writing}). The \emph{answering} verdict dropped from $23/31$ to $15/31$ and mean restated share rose from $17.6\%$ to $25.0\%$. Verbatim fidelity of the generated prompts, judged independently, fell from $19/24$ to $13/24$.

\begin{table}[h]
  \centering
  \caption{Prose versus nested XML for the same instruction content, $n{=}31$ per arm.}
  \small
  \begin{tabular}{lcc}
    \toprule
    Metric & Prose & XML \\
    \midrule
    \emph{answering} verdict & \textbf{23/31} & 15/31 \\
    Mean restated share & \textbf{17.6\%} & 25.0\% \\
    Verbatim fidelity & \textbf{19/24} & 13/24 \\
    Requester-as-subject openings & 4 & 10 \\
    \bottomrule
  \end{tabular}
  \label{tab:writing}
\end{table}

The failure forms locate the mechanism. Classifying the opening sentence the judge quoted in each defective answer, the requester-as-subject form ("\emph{[the requester]} requires a mechanism that\ldots") more than doubled. Nested XML foregrounds the requester as a first-class tagged entity at the top of the most privileged source in the prompt, which makes it the most available grammatical subject. Structure changed what was salient, and salience governs what the model writes first.

Nothing here contradicts the reading results of \S\ref{sec:reading}, or the literature they reproduce: those concern locating things in a document. What the pair marks is the edge of that claim. The distinction the evidence supports is between material a model must \emph{locate things in}, where structure is signal, and material it must \emph{internalise and write from}, where structure redistributes salience.

\subsection{Naming a construction concentrates it}
\label{sec:prohibition}

The defect above -- opening an answer by restating the requirement -- was localised: $38$ of $39$ offending sentences sat in the first $15\%$ of their answer, at median relative position $0\%$. We addressed it with a test the model applies to its own sentence, rather than a list of banned words: write the first sentence, delete everything in it that came from the requirement, and if nothing specific remains, write it again. The two surface forms this suppressed fell from $35/144$ and $37/144$ openings to $2/44$ and $0/44$, and the verdict rose from $5/40$ to $27/40$ on T1.

The final prompt additionally \emph{names} the two remaining traps, describes each, and supplies a worked example of each. Table~\ref{tab:prohibition} classifies the $26$ surviving defects: $25$ ($96\%$) are in exactly those two named forms, and none is in either form suppressed without naming. A fresh run reproduced this at $5/5$.

\begin{table}[h]
  \centering
  \caption{Surviving defects by surface form, and whether the prompt names that form.}
  \small
  \begin{tabular}{lcc}
    \toprule
    Surface form & $n$ & Named? \\
    \midrule
    Nominalised passive & 14 & yes \\
    Requester-as-subject & 11 & yes \\
    Product-name opener & 1 & yes \\
    Modal \emph{must}/\emph{shall} & 0 & yes \\
    \midrule
    Suppressed by the strip test & --- & no \\
    \bottomrule
  \end{tabular}
  \label{tab:prohibition}
\end{table}

The same pattern held in two unrelated interventions. A prose rule explaining that a particular adjective must not be used moved its occurrences $11 \rightarrow 18 \rightarrow 29$ across successive attempts. A notes block titled "remove X across this bid", which named the forbidden token five times in the highest-attention position of the prompt, produced \emph{denials} that named it -- "without requiring an X-hosted environment" -- raising a question the requester had never asked. The one intervention that named no prohibition, wrapping illustrative content in an \texttt{<example>} tag declaring it a shape to imitate rather than text to copy, moved verbatim copying $13 \rightarrow 0$ on the first attempt and has held at zero since.

This is consistent with the finding that constraint \emph{design} dominates constraint \emph{formatting} for compliance \cite{tang2026compact}: rewording the prohibition never worked, and changing what the instruction asked for worked immediately. It also bears on \citet{khan2025inversion}, who report that constrained rule-based prompting outperforms chain-of-thought on a mid-capability model ($97\%$ versus $93\%$) and \emph{underperforms} it on a frontier model ($94\%$ versus $96.36\%$), attributing the reversal to constraints that guard a weaker model becoming handcuffs on a stronger one. Our deployment is open-weights and not frontier-class, which places it on the side of that interaction where heavy structural constraint is the right strategy. The prediction this carries is that the four-turn scaffolding reported here would \emph{degrade} performance if the underlying model were replaced with a frontier one. We have not tested that, and it is the single most falsifiable claim in this paper.

\subsection{Variance amplification through windowing}
\label{sec:variance}

We re-ran the full pipeline on T4 four days after a reference run, on the byte-identical input file (Table~\ref{tab:variance}). Element parsing, past-bid segmentation, diagram conversion, exemplar selection and reference assembly were exactly reproducible. Slot marking was not: $54$ versus $56$ slots.

\begin{table}[h]
  \centering
  \caption{Two runs, one byte-identical $349$-element file.}
  \small
  \begin{tabular}{lcc}
    \toprule
    Quantity & Run 1 & Run 2 \\
    \midrule
    Elements parsed & 349 & 349 \\
    Elements echoed back & 349/349 & 348/349 \\
    Answer slots found & 54 & 56 \\
    Window 2 span & E0104--E0273 & E0107--E0275 \\
    Questions extracted & \textbf{68} & \textbf{51} \\
    \bottomrule
  \end{tabular}
  \label{tab:variance}
\end{table}

The question count moved $68 \rightarrow 51$, and the window spans show why. Question extraction computes its windows over the \emph{marked} element stream; adding an \texttt{answer\_slot} attribute changes an element's rendered character length; so a two-slot difference shifts the window boundaries by three elements, changing what each extraction call sees.

No agent hallucinated and no false claim propagated. Variance amplified because a deterministic length-based function was fed a stochastic annotation. The mitigations proposed for semantic cascades -- boundary gates, clarification modules \cite{lin2025agentask}, provenance graphs \cite{xie2026spark} -- do not apply, because there is nothing false to detect. Any pipeline that windows, chunks or paginates over model-annotated text has this property, and it is invisible unless the boundaries are logged. The remedy is to window over the unmarked stream and apply annotations afterward.

\subsection{Compaction, and what it costs}

Because the model is stateless, the entire message array is posted on every call. On one six-question session the final call carried $355{,}024$ characters, $73\%$ of it plan and compliance text about \emph{other} questions. Collapsing each completed question to (question, committed answer) cut characters posted across eleven sessions from $10.2$M to $2.1$M ($-79\%$), consistent with the upper end of the reduction reported for optimised context compression \cite{kang2025acon}.

The cost is easy to miss: material attached to one question does not survive it. Across all $22$ checkpointed sessions, zero retained turns still carry the past-bid exemplar or the capability-facts block; only the reference block survives, because it is explicitly re-attached to the retained turn. Anything that must reach every answer has to ride the channel compaction preserves, not an individual question's prompt.

\subsection{Corroboration is not validation}
\label{sec:corroboration}

Because judge and author use the same model family, self-preference bias is a material concern; it is perplexity-driven \cite{wataoka2024selfpreference} and is not reliably eliminated by prompting alone. We used lexical counts computed outside the pipeline as a check, and record that this failed.

The counts were at their best -- modal openings $2/44$, specification-voice openings $0/44$ -- in precisely the run whose judged verdict was worst ($14/40$). They were measuring surface forms the intervention had genuinely eliminated while the behaviour relocated into unmeasured forms at the same textual position. Only a metric defined over the sentence's \emph{function} detected the regression. Mechanical corroboration does not rescue a badly specified metric; it makes a confident wrong answer cheaper.

This is also why \S\ref{sec:truth} matters disproportionately. A comparison against work a human wrote and submitted is the one measurement in this paper whose reference point was not produced by the system under test.

\section{Discussion}

The two halves of this paper are connected. A system whose answers are competitive with submitted human work on three quarters of a document (\S\ref{sec:truth}) is not achieving that through model capability alone -- the model is open-weights, non-frontier, and had no worked example on that tender. It is achieving it by being given the requester's own words verbatim, the right structure to read them in, and a set of tests to apply to its own output.

The gap analysis also relocates where the remaining effort should go. Only a fifth of the shortfall against human work is a generation problem; two thirds is knowledge the organisation holds and the pipeline was never handed. That argues for investment in evidence capture -- getting subcontractor registers, prior correspondence and institutional decisions into the corpus -- rather than in further prompt engineering. It also implies a limit on what any benchmark of this kind can tell us: a system evaluated against a human artefact is being scored partly on its inputs, and a study that reports only the aggregate verdict cannot distinguish a better writer from a better-briefed one.

The asymmetry result says where each of those instruments belongs. Structure is what makes reading reliable, and the three reading tasks in \S\ref{sec:reading} moved from unusable to reproducible on that basis alone. But the same instrument applied to instruction material cost $26$ points of answer quality, because it changed which entity was most salient at the top of the most privileged source in the prompt. The prohibition result (\S\ref{sec:prohibition}) is the same lesson in a different register: an instruction that describes a failure supplies the tokens for it, while an instruction that describes a test the model runs on its own sentence does not.

Read together with \citet{khan2025inversion}, this suggests the engineering effort here buys the most where the model is weakest, and should be expected to depreciate as models improve. That is a comfortable conclusion for a research programme and an uncomfortable one for a product.

\section{Conclusion}

We evaluated a deployed multi-agent tender-response system against human bids the same organisation submitted. On a blind comparison in which the system had no worked example, an LLM judge rated it at least as good as the human-submitted answer on $40$ of $55$ ground-truth sections, better on $4$, missing on none, with one unsupported claim flagged in total. Classifying every gap showed $68\%$ to be information absent from the system's sources rather than a writing failure, which moves the score to $49$ of $55$ once those are excluded -- and which argues that comparisons of this kind should report both figures, since the difference between them measures the briefing rather than the writer. Where a human instead post-edited the system's draft, $20.5\%$ of the final human text originated in it.

We then reported a boundary on a well-established result. Rendering documents as structural markup improved every reading task we instrumented -- annotation routing from $61\%$ to $97.8\%$ coverage, workbook slot counts from a $124$-cell swing to a $1$--$2$ cell disagreement, past-bid segmentation from unstable to bit-identical -- and reversed on the one conditioning task we ran as a controlled pair, dropping answer quality from $74\%$ to $48\%$. We also find that naming a forbidden construction concentrates the residual defects into exactly the named forms, and that coupling a stochastic annotation to a deterministic windowing function turns a two-slot difference into a $17$-question one. Structure belongs where the model reads. Where it writes, the effective instrument is a test it applies to its own output.

\section{Limitations}
\label{sec:limitations}

The ground-truth comparison in \S\ref{sec:truth} is a single procurement judged by a single LLM judge from the same model family that produced the drafts, with no blinded human scoring; \S\ref{sec:corroboration} shows our intended mitigation for judge bias did not work, and a debiasing protocol \cite{soumik2026judging} would be the correct next step. The judge's documented preference for markdown formatting is an unmeasured confound in a system whose answers are markdown. We did not attempt the reverse comparison of human answers judged against ours.

The gap classification carries its own circularity: the same model family that wrote the answers and judged them also decided which gaps were unavoidable, and it had an evident interest in the answer. We mitigated this by supplying mechanical evidence of whether each named entity occurs in the corpus, but the final call is still a model's. The classifier returned no \emph{policy divergence} label at all despite our having identified one such case by hand, which suggests it under-detects that category. The honest position is that $73\%$ and $89\%$ bracket the result and neither is a point estimate.

The post-editing measurement in \S\ref{sec:postedit} records what one editor kept, not whether the draft saved time, and $n{=}1$ editor on $n{=}1$ procurement. The conditioning comparison in \S\ref{sec:writing} has $n{=}31$ per arm on a single procurement and has not been replicated; a $26$-point difference at that sample size is suggestive rather than conclusive. Runs were performed in sequence during active development, so several comparisons vary more than one factor; only the paired conditioning comparison and the three-run slot reproducibility hold exactly one factor. The past-bid corpus holds two eligible tenders, so exemplar selection is measured over a two-candidate choice. The variance coupling in \S\ref{sec:variance} is identified but not fixed.

\bibliography{main}

\appendix

\section{The Four-Turn Chain}
\label{sec:appendix}

The model has no thinking tokens, so no XML reasoning envelope is used: the conversation is stateful, so the turn boundary \emph{is} the envelope. A plan turn's reply conditions the draft and no code reads it as an answer, which also means a tender containing a closing tag cannot break the format. Median prompt lengths over T4's $44$ answers: plan $2{,}188$ characters, draft $12{,}114$, compliance $8{,}693$, quality $10{,}631$.

The plan turn is a self-ask over seven questions rather than a free monologue. We observed monologue diversity collapse directly: as answers grew $22\%$ longer, distinct opening words fell $86 \rightarrow 46 \rightarrow 30$. One question asks what the requirement demands that the model would not do by habit, because tenders are full of counter-conventional instructions -- a page limit, a required order, a required form of words -- that trained habits override unless the model is made to notice them first. Another requires every figure, volume, rate, date and named system in the question to be listed before any reasoning, which addressed a cluster of arithmetic defects in pricing cells.

The quality turn may not delete. Its first version could remove material it judged unsubstantiated, and did: on one bid it cut $40$--$75\%$ of the strongest evidence, because a named language or a measured result carries no architecture or standard and so reads as unsupported abstraction to an instruction to strip them. Its permitted moves are now to add, reframe, flag, or mark text for a human to decide on, and the runner independently reports any turn that shortens an answer by more than $15\%$.

\section{Ground-Truth Protocol}
\label{sec:appendix-truth}

The human submission for T3 was recovered as $55$ (question, answer) sections by the same LLM-over-markup pass the system uses to read a past bid, then supplied to the validator as ground truth. We verified independence three ways before treating it as blind.

First, provenance: the human response was written and submitted before this system existed. Second, self-exclusion: $0$ of $120$ generated T3 prompts contained an exemplar block, so no part of the human response reached the drafter. Third, textual: excluding $10$-grams that also appear in the blank forms, the human's own answer text and the system's share \textbf{$8$ $10$-grams} in total -- $0.0\%$ of the system's answer text. Those eight are shared source facts drawn independently from the same internal collateral, such as a token price and a data-handling policy, which is the convergence one should expect and not evidence of contamination.

The same test disqualified T4 as ground truth, and it is worth recording how close that call looked at document level. Comparing the two filled documents directly suggested $38.8\%$ overlap, which would have been alarming; $26\%$ of the human document is the blank form's own text, and once that boilerplate is excluded the reuse is $20.5\%$ of the human's answer text. The document-level figure was mostly the requester's own questions appearing in both files. Any post-edit study on form-based documents needs the blank form as a third term, or it will misattribute boilerplate to reuse.

\paragraph{Gap classification.} To separate information-availability failures from writing failures we reconstructed the system's corpus for T3 exactly as the run had assembled it: each reference source truncated at its configured cap, plus the tender's own form and four addenda, giving $372{,}139$ characters. For each gap we extracted its capitalised multi-word entities and recorded, mechanically, which occur anywhere in that corpus and which do not. Gap text, present entities and absent entities were then passed in batches to a classifier asked to assign exactly one of \emph{information unavailable}, \emph{information available and unused}, \emph{policy divergence} or \emph{additional detail}, with the instruction that a system can only write what its sources support. All $70$ gaps were classified; two records whose gap text was empty were discarded before analysis.

The design is deliberately two-part -- a mechanical presence test, which is a question of fact, feeding a semantic judgement, which is not. It remains circular in the sense noted in \S\ref{sec:limitations}, and the classifier's failure to return any \emph{policy divergence} label is direct evidence of that circularity rather than evidence the category is empty.

\end{document}